\documentclass[sigconf]{acmart}

\AtBeginDocument{%
  }

\setcopyright{acmlicensed}
\copyrightyear{2018}
\acmYear{2018}
\acmDOI{XXXXXXX.XXXXXXX}

\acmConference[Conference DIS'25]{Designing Interactive Systems}{July 05--09,
  2025}{Funchal, Madeira}
\acmISBN{978-1-4503-XXXX-X/18/06}

\usepackage{xcolor}
\usepackage{bbm}

\begin{document}

\title{ELEGNT: Expressive and Functional Movement Design for Non-anthropomorphic Robot}




\author{Yuhan Hu}
\affiliation{%
  \institution{Apple}
  \city{Cupertino}
  \country{United States}}
\email{yhu58@apple.com}

\author{Peide Huang}
\affiliation{%
  \institution{Apple}
  \city{Cupertino}
  \country{United States}}
  \email{peide_huang@apple.com}

\author{Mouli Sivapurapu}
\affiliation{%
  \institution{Apple}
  \city{Cupertino}
  \country{United States}}
\email{mouli.siva@apple.com}

\author{Jian Zhang}
\affiliation{%
  \institution{Apple}
  \city{Cupertino}
  \country{United States}}
\email{jianz@apple.com}




\renewcommand{\shortauthors}{}

\begin{abstract}
Nonverbal behaviors such as posture, gestures, and gaze are essential for conveying internal states, both consciously and unconsciously, in human interaction. For robots to interact more naturally with humans, robot movement design should likewise integrate expressive qualities—such as intention, attention, and emotions—alongside traditional functional considerations like task fulfillment, spatial constraints, and time efficiency. In this paper, we present the design and prototyping of a lamp-like robot that explores the interplay between functional and expressive objectives in movement design. Using a research-through-design methodology, we document the hardware design process, define expressive movement primitives, and outline a set of interaction scenario storyboards. We propose a framework that incorporates both functional and expressive utilities during movement generation, and implement the robot behavior sequences in different function- and social- oriented tasks. Through a user study comparing expression-driven versus function-driven movements across six task scenarios, our findings indicate that expression-driven movements significantly enhance user engagement and perceived robot qualities. This effect is especially pronounced in social-oriented tasks. 

\end{abstract}


\begin{teaserfigure}
\centering
  \includegraphics[width=\textwidth]{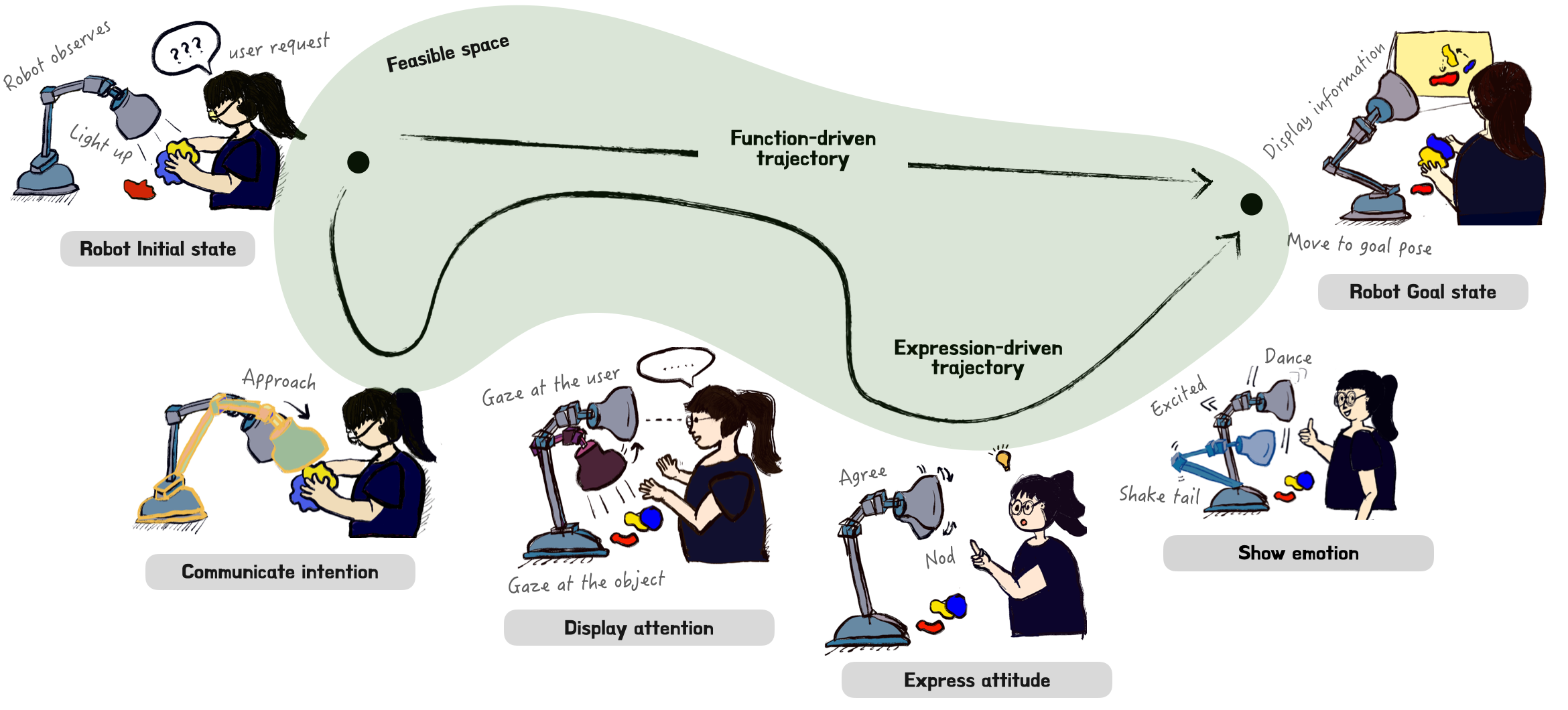}
  \caption{Overview of our research hypothesis: robots should not only move to fulfill functional goals and constraints, i.e., robot moving from the initial state to goal state through a shortest, feasible trajectory (function-driven trajectory), but also use movements to express its internal states to human counterparts during the interaction, i.e., via expression-driven trajectory to express robot's intention, attention, attitude, and emotions.}
  \Description{}
  \label{fig:teaser}
\end{teaserfigure}


\maketitle

\section{Introduction}


In this paper, we present \textbf{ELEGNT}, a framework of \textbf{\underline{e}}xpressive and functiona\textbf{\underline{l}} mov\textbf{\underline{e}}ment desi\textbf{\underline{g}}n for \textbf{\underline{n}}on-anthropomorphic robo\textbf{\underline{t}}. 
We argue that robots should not only move to fulfill functional purposes and constraints but also move ``elegantly'' - using their movements to express intentions, attention, and emotions to their human counterparts during human-robot interactions (HRI). We present our practice of designing movements incorporating functional and expressive utilities and a user research to understand the effect of expressive movements.

Robots are increasingly entering homes as assistants and companions, making it essential to understand how they coexist with humans, interact with people, and fulfill functional and social roles in everyday life. Like most animals, humans are highly sensitive to motion and subtle changes in movement. Existing research on robotics suggests that a robot's movements can not only perform practical functions but also convey the robot's purpose, intent, state, character, attention, and capabilities \cite{hoffman2014designing}. 

While much of the research separates pragmatic robotics—such as robotic arms performing household tasks—from social robotics, like therapeutic robots providing emotional support, we argue that any robot interacting with humans, even if designed primarily for practical functions, embodies social value and should have its design and behaviors shaped accordingly. For instance, in a collaborative manipulation task with a human teammate, a robot should not only consider functional actions, such as picking up and placing objects, but also employ expressive movements that convey its intentions, state, and even character traits. These expressive cues can help human collaborators anticipate the robot’s actions, build trust, and foster a sense of comfort and enjoyment in the collaborative process.

Our research addresses several questions: How do we design expressive movements alongside functional actions for robots interacting with humans? What are the design spaces and movement primitives? How do users perceive robots employing expressive movements versus purely functional ones? 

In this paper, we present our practice on designing an non-anthropomorphic robot in the form of a lamp, featuring a 6-DOF arm and a head equipped with a light and a projector. 
As a common household form factor, the lamp-like robot offers a rich design and interaction space to engage with both the environment and users through lights and movements—for example, directing a user’s attention by illuminating specific spaces or objects. 

We use research-through-design (RtD) \cite{gaver2012should} methodology to iterate the design of the robot's form, movement, and interaction scenarios. 
We formulate movement objectives with both functional and expressive utilities - whereas functional utility brings the robot's from initial to goal states within the physical and task space, expressive utility emphasizes the trajectories taken to achieve these goals. The latter incorporates considerations for expressing and communicating the robot's intention, attention, attitude, and emotional state to the user, as illustrated in figure \ref{fig:teaser}.
We elaborate the building blocks for expressive movements with kinesics and proxemics primitives. 
Through video prototyping and storyboarding, we demonstrate a range of use cases and task scenarios for the lamp robot in home environments, organized along the dimensions of robot agency and the social versus functional nature of tasks.
Our work aims to offer design inspiration and a framework for future integration of expressive robots into daily life.

To evaluate the benefits of incorporating expressive movements and comparing the outcome between expressive and functional utilities, we conducted a user study comparing expression-driven movements with function-driven movements across various task scenarios. Participants (n=21) were tasked to watch human-robot interaction videos in six different tasks, each with two robot variations. After each video, they evaluated the perception along the metrics of engagement, intelligence, human-likeness, willingness to interact, sense of connection, and robot character.

The results reveal that movements incorporating expressive utilities significantly increased user ratings, compared to movements only driven by functions. 
Perceptions varied across tasks, with expressive movements particularly benefiting social-oriented tasks, such as entertainment and social conversations.
Results also suggested demographic effects from participants' age and professional backgrounds.
Qualitative analysis revealed additional insights into perceived robot characteristics and how users infer the robot’s state from its movements, suggesting the potential to customize these movements to individual preferences and align them with other interaction modalities, including voice and light.
We hope the proposed framework and our study outcomes will inspire future research on expressive movement generation for non-anthropomorphic robots.

\section{Related Work}

In this section, we review related work in the domain of non-anthropomorphic robots for human-robot interactions, with an emphasis on robot expression and movement-centric design. Additionally, we discussed related work from animation and character design, which highly inspired this work.

\subsection{Non-anthropomorphic Robots for HRI}

\begin{figure}[t]
    \centering
    \includegraphics[width=1\linewidth]{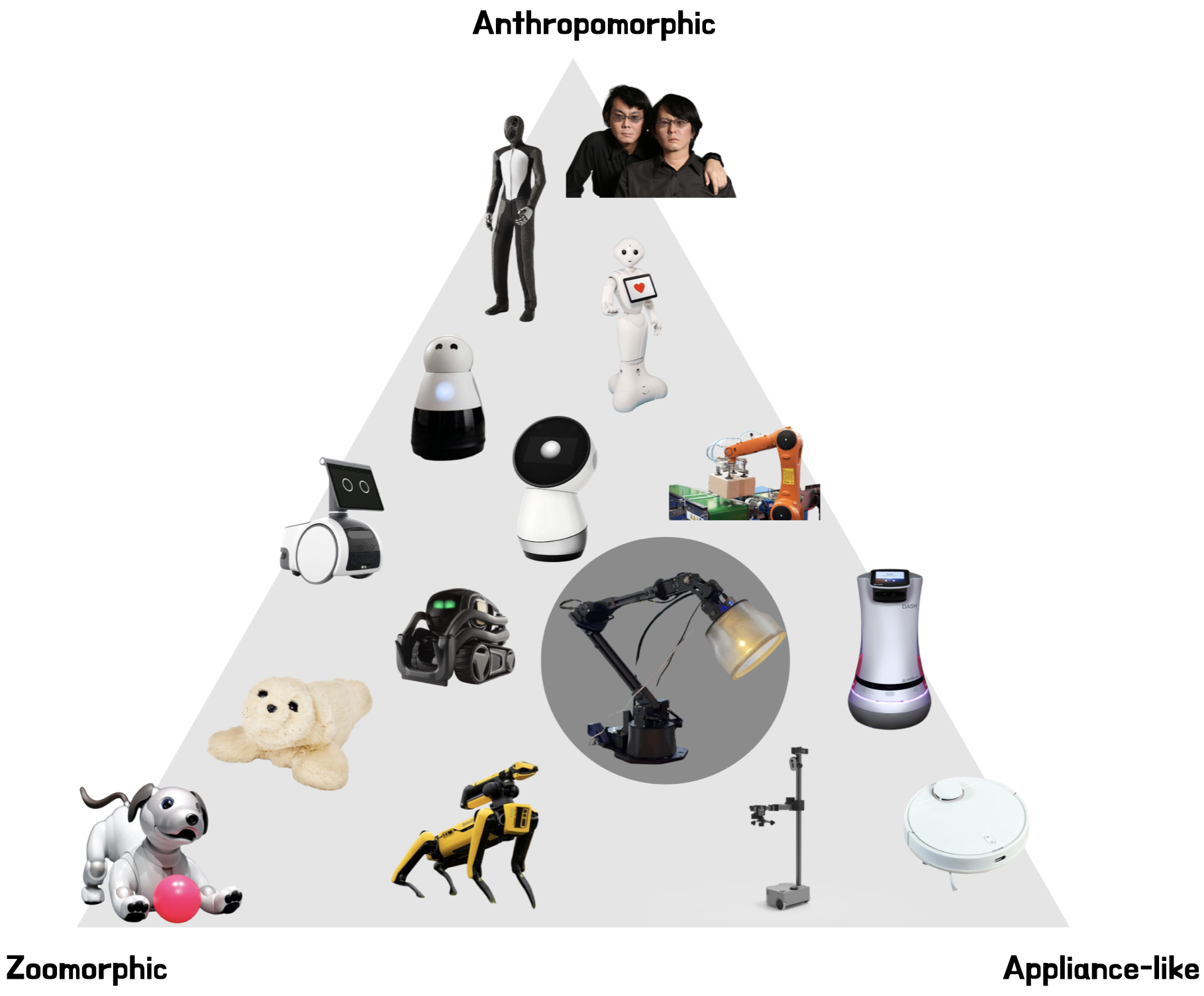}
    \caption{Existing robot form spectrum: Anthropomorphic, Zoomorphic, and Appliance-like.}
    \label{fig:anthropomorphism}
\end{figure}

The form and appearance of a robot impacts how people perceive it, interact with it, and build long-term relationships with it \cite{bartneck2004shaping}.
Existing robotic form can be categorized with anthropomorphic (human-like) \cite{huang2024emotion}, zoomorphic (animal-like), and appliance-like, as illustrated in figure \ref{fig:anthropomorphism}.
While it is beneficial to have robots with anthropomorphic design that have a positive impact on acceptance \cite{duffy2003anthropomorphism}, research also suggested that user preferences of robot forms were task and context dependent \cite{goetz2003matching}. The appearance of robots should match its capabilities and user expectations. 

Anthropomorphic robots use human-like gestures and expressions, such as body pose and facial expressions, mapped from humans' behaviors to convey a various of internal states. Non-anthropomorphic robots does not have an explicit vocabulary or mapping for expression.
Existing research suggest multiple expressive channels for non-anthropomorphic robots, including movements \cite{hoffman2014designing, luria2016designing, bretan2015emotionally, koike2023exploring, takayama2011expressing}, light/color \cite{loffler2018multimodal, luria2017comparing}, sound \cite{kozima2009keepon}, tactile expression \cite{hu2019using, bucci2017sketching, hu2023can}, and so on.
For instance, Shimon \cite{hoffman2010shimon}, a musical improvisation robot, incorporates a socially expressive head to communicate its internal states including rhythm, emotional content, intensity, as well as manage turn-taking and attention between the robot and human musicians, supporting joint musical attention.
\textit{``The Greeting Machine''} \cite{anderson2018greeting}, embodies a small ball rolling on a larger dome, designed to communicate positive and negative social cues in the context of opening encounters. 
Existing studies suggest that even abstract and simplistic movements, such as \textit{Approach and Avoid}, are effective in expressing robot intentions, evoke positive and negative experiences of the users. 

This work is highly inspired by the paper \textit{''Designing Robots with Movement in Mind''} \cite{hoffman2014designing}, which present technics of movement centric design, including character sketches, video prototyping, and Wizard of Oz studies. They illustrate the approach and design strategy with the design of non-anthropomorphic robots and robotic objects.
They suggest movements as robot dynamic affordance, which help cues users on potential actions and interactions that the robot is capable of. The robot's expressive movements were considered early on in the design process, and may co-evolve in the design iterations with robot hardware appearance and use cases. 
Many recent research \cite{gemeinboeck2017movement, kahn2008design} shift the robot design focus from the production of life-like forms to the process of movement and kinesthetic creation. 
It is important to consider the expressive power of the robot's movement - design movements that express the robot's purpose, intent, state, mood, personality, attention, responsiveness, intelligence, and capabilities \cite{nakata1998expression}. 
In this work, we take the idea of movement-centric design further with illustrated movement designs grounded in real-world interaction scenarios and conduct a user study to evaluate the effects of expression-driven movements versus purely functional ones.
Additionally, our work follows the common practices of research-through-design approaches, combining artefact-centered research \cite{cila2021learning} and speculative design explorations \cite{alves2021collection, hu2024designing}.



\subsection{Movement Design for Expression}

Movement plays a fundamental role in how humans perceive and interact with the world. Humans, like many animals, are highly sensitive to motions \cite{hall1966hidden}. Movements are essential in the coordination and performance of joint activities, serving to communicate intentions and refer to objects of shared attention \cite{clark2005coordinating}.

Insights into expressive movement design can be drawn from domains beyond robotics, such as animation, behavioral science, and performing arts \cite{loke2013moving}. In these fields, movement is used as a medium for communication, enabling objects, characters, and forms to convey emotions, intentions, and narratives. For example, in character animation, abstract forms like dots, lines, and shapes are brought to life through motion, timing, and staging. A classic example is the animated short \textit{"The Dot and the Line"} \cite{thedot}, where all expressions are conveyed through motion with minimal visual elements. The Pixar's iconic animation \textit{"Luxo Jr."} \cite{luxo}, also serves as the primary inspiration of the lamp form, featuring two desk lamp characters, demonstrating how simple movements can communicate narratives, relationships, and emotions. These works highlight the power of motion, even with simple geometry, is effective in storytelling and expression.

Research also shows that movements do not need to mimic human forms in detail to be perceived as intentional or expressive. Humans are adept at interpreting the movements of abstract shapes, as demonstrated by studies on point-light displays \cite{kozlowski1977recognizing}, where participants could classify activities and recognize individuals from minimal visual cues. Beyond recognition, humans often attribute internal states, characters, and intentions to abstract movements, as exemplified by the \textit{Heider-Simmel Illusion} \cite{abu2011neuroanatomical}, where simple geometric shapes moving in suggestive ways are perceived as having purpose or personality. This phenomenon is closely tied to the Theory of Mind \cite{baron1991precursors}, which describes humans' ability to infer mental states and intentions from observed behaviors.

Drawing inspiration from animation principles and leveraging humans’ innate sensitivity and projection to motion, we aim to design and program physical robots with movements that effectively convey expressive and intentional behaviors. These principles form the foundation of our work, combining the expressiveness of motion with functional considerations to create engaging and meaningful interactions.

\begin{figure}[t]
    \centering
    \includegraphics[width=0.95\linewidth]{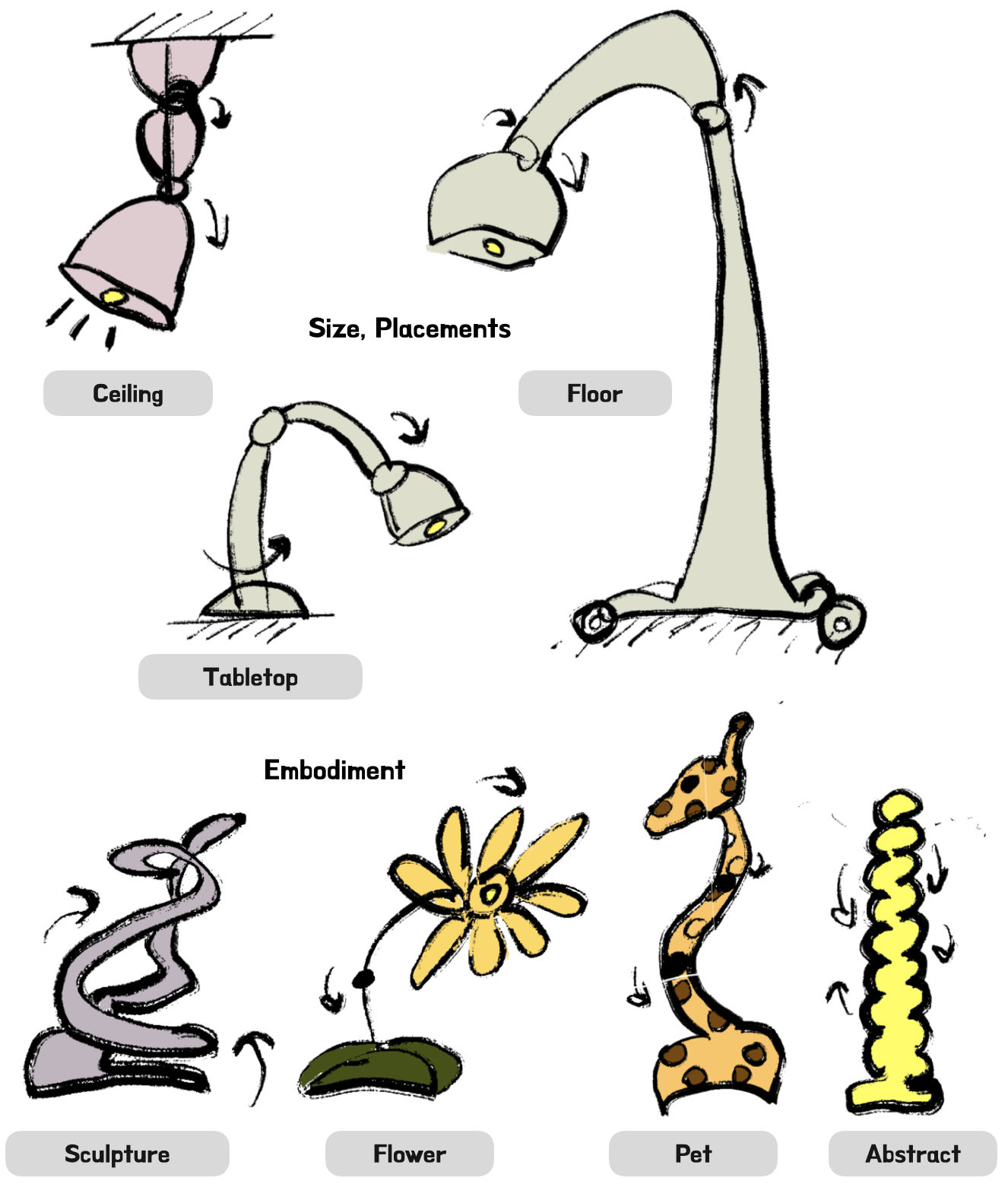}
    \caption{Sketching ideas of non-anthropomorphic robots with different form factors, sizes, and placements.}
    \label{fig:form_sketch}
\end{figure}

\section{Methodology}

In this section, we outline the design process of a lamp-like robot, developed by a team of HRI researchers, roboticists, and animators. 
Through iterative brainstorming, sketching, storyboarding, and both hardware and software prototyping, we explored a range of design considerations, including form selection, movement design, and potential use cases. 
Rather than providing an exhaustive taxonomy of design spaces, our aim is to highlight key design opportunities and primitives that can inspire and guide future research and practice.


\subsection{Designing Robot Form}

There have been numerous explorations into home robots, such as vacuum robots \cite{tribelhorn2007evaluating}, table-top robotic assistants \cite{rane2014study}, robot pets \cite{fujita2004activating}, and humanoids \cite{kaneko2002design}.
These robots often take on anthropomorphic, zoomorphic, or appliance-like forms, as illustrated in figure \ref{fig:anthropomorphism}. 
Existing HRI research indicates that a robot’s form can shape user expectations and influence interaction affordance. For instance, users may expect a humanoid robot to interpret facial expressions and gestures, whereas a vacuum robot might invite less social engagement. Aligning a robot’s form with user expectations and functional capabilities is a critical design consideration.

Inspired by the characters in \textit{``Luxo Jr.''} \cite{luxo}, we adopt the form factor of a desk lamp. Although primarily appliance-like, it incorporates subtle anthropomorphic elements—such as the lamp head and the arm connecting the head to the stand—that evoke the appearance of a head and neck. 
The lamp’s light and camera can also be mapped to robot ``eyes'', providing a design opportunity to convey the robot’s attention and purpose.

\begin{figure*}[t]
    \includegraphics[width=0.85\textwidth]{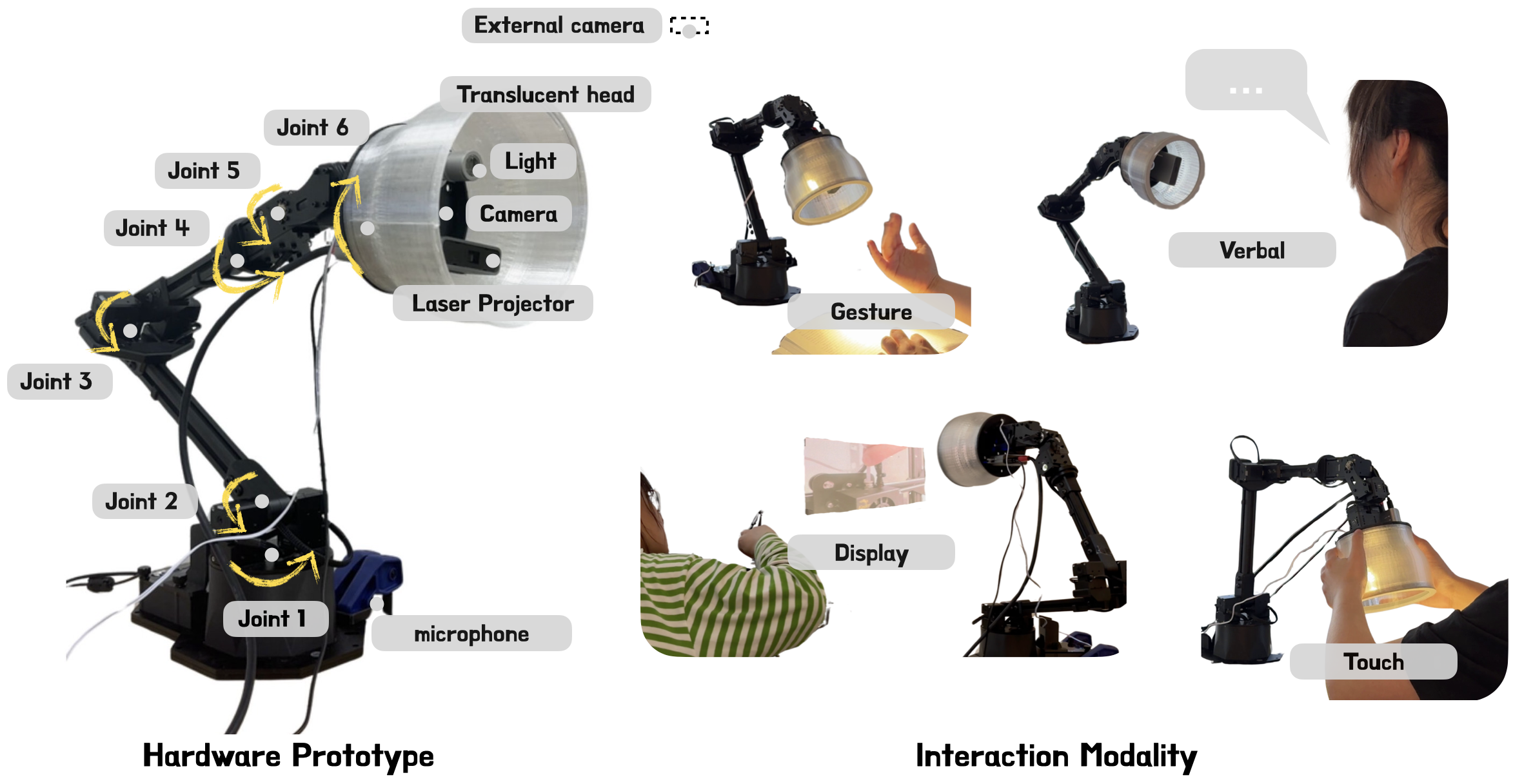}
     \caption{Hardware composition of the lamp robot (left); Interaction modalities between human and robot, including gesturing, verbal communication, light and projection display, and touch interaction (right).}
     \label{fig:hardware}
\end{figure*}

We consider aesthetic, expressive, and pragmatic aspects when prototyping the hardware of the robot.
From a pragmatic perspective, we aim for the robot to have a wide range of motion, allowing it to cover a reasonable interaction space—for example, transitioning from illuminating a table to lighting up a couch. We explored various placements and configurations for the lamp robot, including ceiling-mounted, tabletop, and floor lamp designs, as illustrated in Figure~\ref{fig:form_sketch}. While a floor lamp offers broader spatial coverage and the potential for mobility, it also introduces challenges such as increased control complexity and a higher risk of physical collisions during interactions. 
In contrast, ceiling-mounted lamps minimize these risks but are limited in their interaction capabilities, as they can only provide light from a top-down angle. 
Additionally, ceiling robot movements are tend to be neglected during interactions since they fall outside the user’s line of sight.
Beyond spatial coverage, we aim for the robot to use kinesthetic movements for expressive purposes, such as nodding, shaking its head, or leaning forward and backward. Therefore, the motors must be strategically positioned to accommodate these motion ranges.

We also explored other non-anthropomorphic forms with similar range of motions, including designs inspired by a flower and a giraffe, and abstract forms like sculptures and art pieces, as shown in figure~\ref{fig:form_sketch}. 
While this paper focuses primarily on the tabletop lamp integration, we envision that certain design principles may translate across different embodiments. For example, movement speed, pauses, and proxemic relationships could be applied universally. 
However, some design patterns, like a nodding gesture, may map differently depending on the embodiment. In forms with a clear head-neck relationship, the gesture might be immediately recognizable, whereas in more abstract forms, it could invite broader interpretations.

Through iterative rapid prototyping, we integrate a robotic hardware platform for further testing and deployment, as illustrated in figure \ref{fig:hardware} (left). 
The robot is composed of a re-purposed 6-DOF robotic arm \cite{widowx}, a 3D-printed plastic lamp head with an embedded LED light, a laser projector, and an internal camera, plus a downward-facing external camera. Additionally, it is equipped with a voice system to listen and speak to the user.

\subsection{Generating Robot Movements}

Existing research has shown that when communicating stories and evoking emotions, movement often plays a more important role than form. Characters and emotions can be conveyed through the timing and quality of a movement. For example, the \textit{Heider-Simmel Illusion} \cite{abu2011neuroanatomical} demonstrates that even simple geometric shapes can be perceived as human figures if they appear to move on their own accord. 
Humans naturally project character and metaphorical states onto moving objects. By intentionally designing these movements, we can instill character perceptions into robots, creating social bonding and tolerance between robot and human, and make the interactions more enjoyable.

\subsubsection{Framework Formulation}
To consider the problem of generating movements considering both functional and expressive objectives, we present the high-level formulation of the problem to guide the low-level trajectory design and integration.
We mathematically formulate the robot movement problem as a Markov Decision Process (MDP) defined by a tuple $(S, A, P, R)$. At timestep $t \in 0, 1, \ldots, T$, the state  $s_t \in S$ consists of the robot joint angles, the tool states, and the environment states. 
For example, the tool states would include turning the light on and off, as well as projecting images. The environment states include the perceived state of the users and other object of interests in the environment.
The action $a_t \in A$ consists of the change in joint angles and the tool event. 
The transition function then defines $s_{t+1} = P(s_t, a_t)$. For simplicity, we also denote the trajectory $\tau = (s_0, s_1, \ldots, s_T)$. The reward function $R$ consists of two parts: functional utility $F$ and expressive utility $E$.


\textit{Functional Utility} $F$ defines the function-driven utility of reaching certain states:
\begin{equation}
    F(\tau) = \sum_{t=0}^{T} f(s_t)
\end{equation}
Without loss of generality, we assume there is only one goal state $s_g$. In this case, 
$f(s_t) = \mathbbm{1} (s_t = s_g)$
, where $\mathbbm{1}(\cdot$) is the indicator function.

\textit{Expressive Utility} $E$ defines the expression-oriented utility of reaching certain states:
\begin{equation}
    E(\tau) = \sum_{t=0}^{T} e(s_t)
\end{equation}
In this work, we draw on design research methods to define $e(\cdot)$ along the expressive dimensions of attention, emotion, intention, and attitudes, as discussed in Section \ref{sec:movement}.

Finally, the objective is to maximize the total utility:
\begin{equation}
    \max_{a_0, \ldots, a_{T-1}}  F(\tau) + \gamma E(\tau)
\label{eq:utility}
\end{equation}
where $\gamma$ is the weight for the expressive utility, which could vary with different task and user. 
In Section \ref{sec:study}, we present a user study to evaluate the perception difference between the robot movements when taking $\gamma > 0$ versus when $\gamma = 0$.


\subsubsection{Functional and Expressive Utility}
\label{sec:movement}

In the context of the lamp robot, \textit{Functional utility $F$} drives motions that aim at achieving a physical goal state, such as taking the initial state of user's reading activity or an explicit verbal request, the robot moves to face the book and turns the light on, as well as projects assistive information such as a visualization of a content in the book. 
The functional utility is measured based on the level of completion of the task in the goal state, such as whether it moved to the desired position, turned on the light, and projected the accurate information.

\textit{Expressive utility $E$}, on the other hand, motivates the actions aimed at communicating the robot's traits, states, and intents to its human interaction partners. 
For example, the robot may increase expressive utility by looking toward a book before moving to it or displaying curiosity through head tilts. Expressive utility can be measured by users’ perceptions of the robot, including perceived intelligence, intuitiveness, interaction quality, trust, engagement, sense of connection, and willingness to use the robot.
Drawing on Theory of Mind (ToM)—the human cognitive ability to attribute mental states such as beliefs, desires, emotions, and intentions to others—we incorporate the following expression categories to capture expressive utilities in the design of our expressive motion library.

\paragraph{Intention} Intention refers to the purpose behind the robot’s actions and the anticipation of its upcoming movements. For instance, when a robot extends its hand, the user can identify which object the robot intends to pick up and what it plans to do with it, enabling cooperation, supervision, or intervention as needed. In the case of a lamp robot, it might briefly turn its head toward a target before moving to reach or interact with it. This behavior signals the robot’s intention, indicates a shift in attention, and cues the user about the next action.

\paragraph{Attention}
Attention refers to where the robot’s focus is directed, with gaze serving as a strong indicator of that focus. For instance, when a robot looks at an object, it may be analyzing it or preparing for upcoming actions. In the context of a lamp robot—where camera and light act as metaphoric ``eyes''—we design gaze behaviors such that looking toward the user can signal attention, for example when the user is speaking. Similarly, a robot can exhibit joint attention by gazing at or illuminating the same object or event as the user. For example, while a user operates an object, the robot might watch the user’s hand and the object being manipulated.

\paragraph{Attitude}
Attitude refers to the robot’s stance toward a person, object, or event. 
For instance, the robot may express agreement or disagreement through motions such as nodding or shaking its head. 
It can also convey attitudes or confidence in response to an instruction or its own action by varying its movement profile—for example, pausing to indicate hesitation or moving quickly and decisively to show confidence.

\paragraph{Emotion}

While robots do not experience emotions as humans do, their ability to simulate emotional expressions is crucial for creating intuitive, engaging interactions. For instance, a robot might use light, bouncy movements to convey happiness, slow movements to suggest a relaxed state, lower its head to indicate sadness, or employ sudden, jerky motions to signal fear or other negative emotions.

\begin{figure}[t]
    \includegraphics[width=0.5\textwidth]{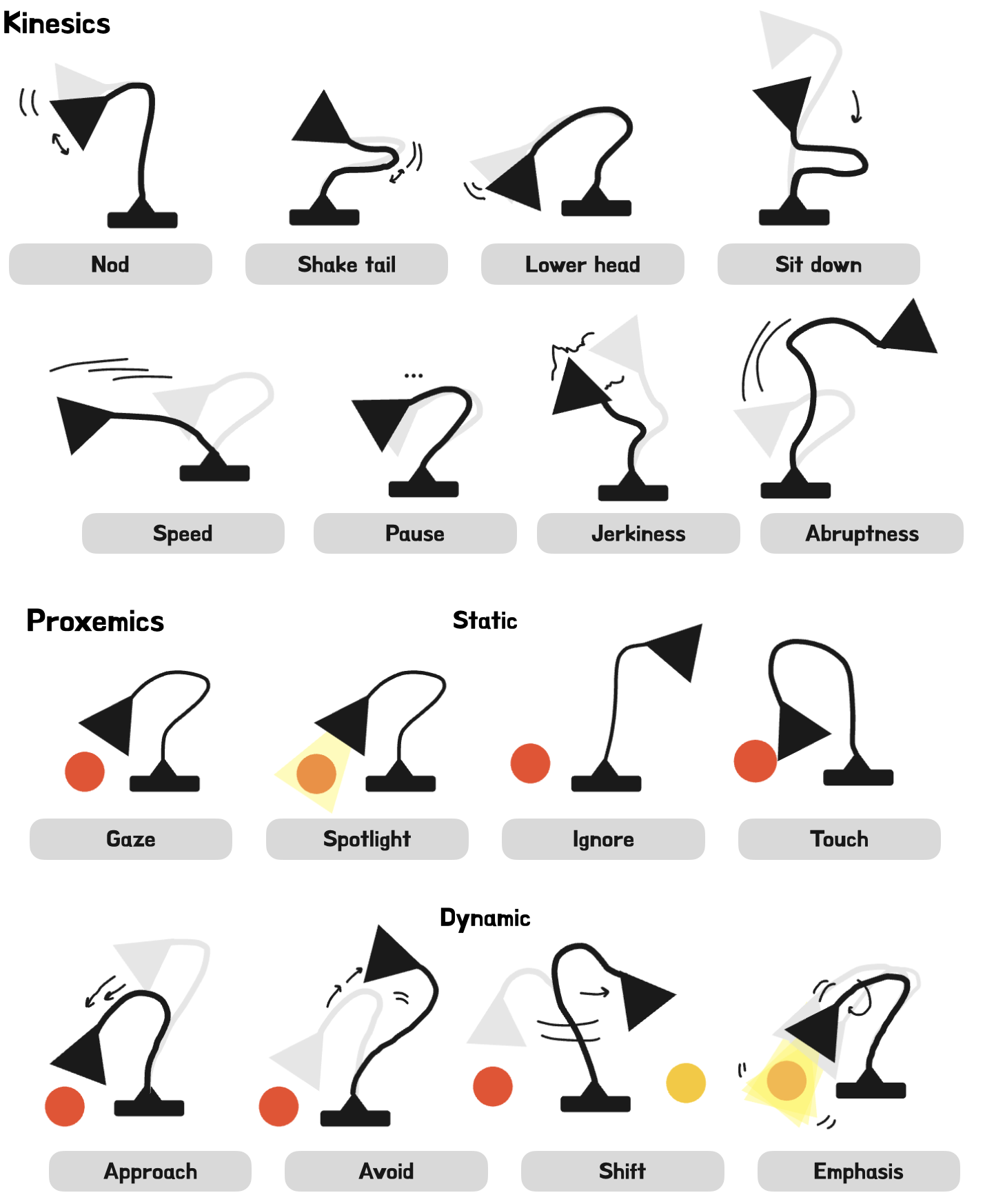}
     \caption{Illustration of the design space for expressive robot movements, including kinesics and proxemics movement primitives.}
     \label{fig:design_space}
\end{figure}

\subsubsection{Building Blocks: Expressive Movement Design Primitives}
\label{sec:primitives}
Through a collaborative effort by animators and robot designers, we developed a design space with several primitives for creating expressive motions, as illustrated in Figure \ref{fig:design_space}. Drawing inspiration from human and animal nonverbal behaviors, we designed motions to express intention, attention, attitudes, and emotions.
Similar to humans, robots can use \textbf{kinesics}—expressive body movements—to communicate information and convey mental states or attitudes \cite{argyle2013bodily}. Kinesics encompass both spatial (pose-related) and temporal features as design primitives.
For \textit{spatial} features, robots can incorporate metaphorical gestures to represent various states. For instance, a lamp-like robot with a head-and-neck configuration might nod or shake its ``head'' to display attitudes, or lower it to convey sadness. The lamp’s long arm joint could also be imagined as a lower body, enabling gestures like ``tail wagging'' to signify excitement or ``sitting down'' to imply relaxation.
For \textit{temporal} features, robots can adjust parameters such as speed, pauses, and acceleration (or jerkiness) to communicate attitudes and emotions. For example, adding pauses and jerky movements might suggest hesitation or a lack of confidence. Varying movement speed can signal different levels of emotional arousal: quick, sharp movements may indicate high-arousal states like excitement or fear, while slower, smoother motions might convey calmness or sadness.

\begin{figure*}[t]
    \includegraphics[width=\textwidth]{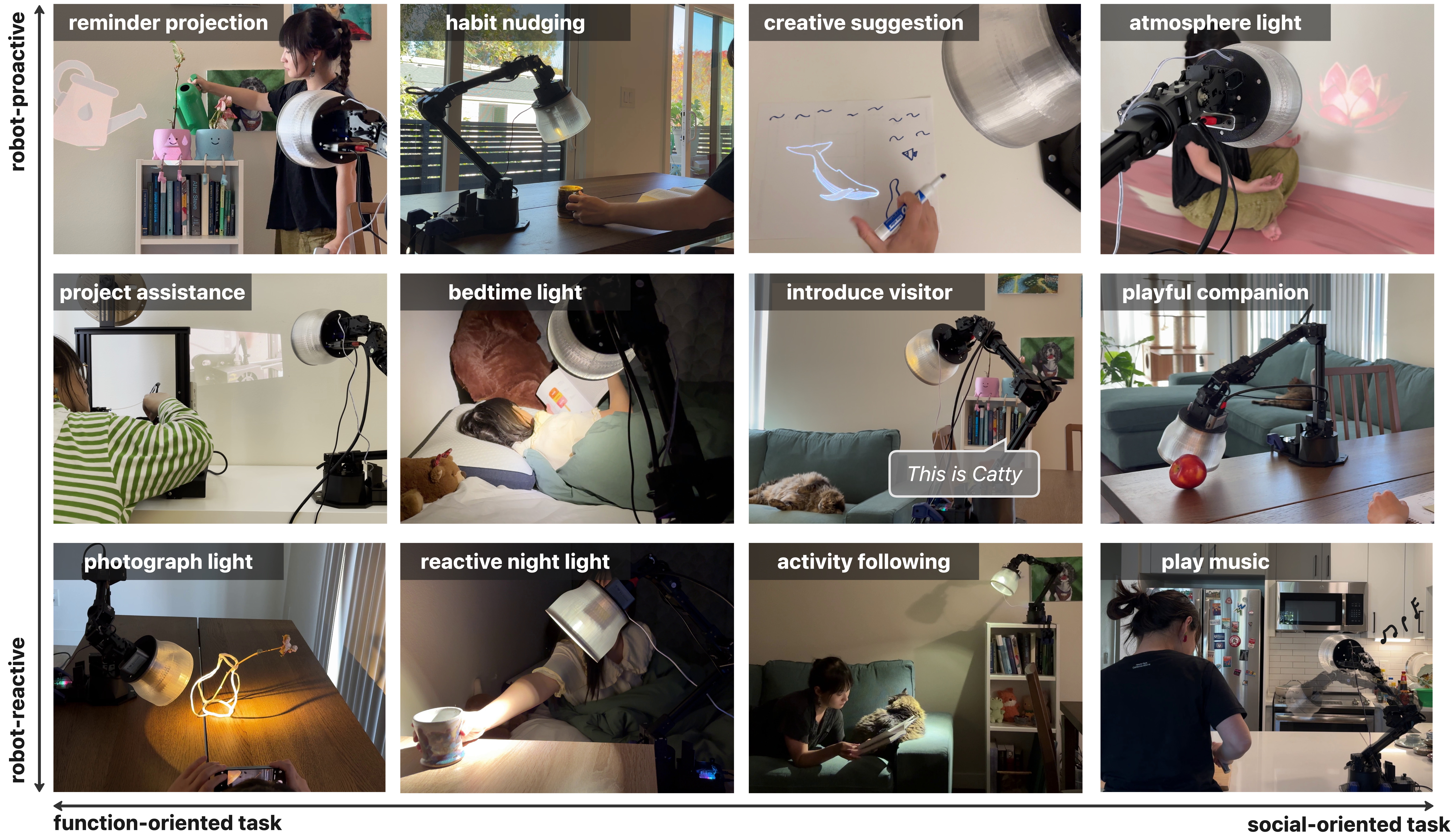}
     \caption{Illustration of at-home interaction scenarios, organized by the robot’s agency (proactive vs. reactive) and task context (function-oriented vs. social-oriented).}
     \label{fig:scenarios}
\end{figure*}

Similar to humans, robots can use \textbf{proxemics}—the management of spatial distance—to express their relationship with the environment and the people around them. This helps set expectations, establish communication channels, create boundaries, and signal context. Proxemics in robots can involve both static and dynamic motion primitives.
For \textit{static} primitives, robots could position themselves relative to an object or person to convey attention and intention. For example, directing their gaze at an object and using light or projection to highlight it can signal focus or communicate context. Pointing their ``head'' away from an object might suggest ignorance or disinterest. Close proximity—such as touching an object—can indicate affection or interest.
For \textit{dynamic} primitives, robots can use movement to express attitudes or intentions. Approaching or avoiding an object may reflect a robot’s stance toward it, while shifting direction between objects or events can indicate changing attention. Dynamic behaviors can also incorporate the use of light accompanied with movements to guide user attention or emphasize a point, such as for reminders or persuasive cues.



In sketching out this design space, we aimed to illustrate how kinesics and proxemics can serve as motion primitives for generating robot expressions. Rather than offering an exhaustive list of design parameters or options, this framework is intended to inspire and guide further exploration and idea generation.





    


\subsection{Interaction Scenarios}

By design, the primary function of the lamp robot is to illuminate spaces and support user activities. Equipped with a projector, the robot can extend this functionality by creating in-situ projections on walls, desks, and other surfaces. This capability enables the robot to project assistive information—either at the user’s request or proactively—to remind or support ongoing activities. For instance, it could project a tutorial video to guide a task or display a creative drawing for inspiration. By projecting onto objects in the environment, the robot can also convey intention or offer contextual information, such as displaying a water icon near a plant to remind the user to water it.

We envision the lamp robot engaging in both social- and function-oriented tasks. Figure \ref{fig:scenarios} illustrates some design outcomes of interaction scenarios and task designs through iterative storyboarding and video prototyping.
On the x-axis, we consider the primary goal of the human-robot interaction. In \textbf{function-oriented} tasks, the lamp robot serves as an assistant or tool—providing information displays, offering desired lighting for user activities, adapting bedtime lighting, and reminding users of schedules or activities. In contrast, \textbf{social-oriented} tasks position the lamp robot more like a friend or pet, emphasizing companionship and entertainment. Examples include suggesting creative ideas, introducing the room to visitors, engaging in playful social interactions, playing music, and projecting atmospheric lighting to enhance the overall user experience.

The second dimension (y-axis) reflects the robot's agency in human-robot interaction, distinguishing between proactive and reactive roles depending on the task.
In \textbf{robot-proactive} tasks, the robot initiates the interaction. Examples include sending reminders, nudging the user to build habits, or offering creative suggestions.
In \textbf{robot-reactive} tasks, the robot responds to user requests or actions. For instance, in a photography lighting task, the robot activates the light based on the user's verbal instructions and adjusts its position in response to pointing gestures. Similarly, a sleep light might switch on or off in response to the user's movements or verbal commands—activating a nightlight when asked or upon detecting that the user has gotten out of bed.

To accommodate a wide range of tasks, the robot employs multiple modalities and activates different input/output channels and skills according to the task requirements. A high-level task manager interprets the lamp’s initial placement, the environment, and contextual information to determine and activate the appropriate state spaces during initialization.
Figure~\ref{fig:hardware} (right) illustrates the various modalities the robot may respond to, including user activities and instructive gestures, speech commands, and touch interactions. The robot leverages torque sensing in its joints and can potentially integrate touch sensors on its surface, enabling it to detect tactile input and switch to compliant modes when needed.

Through the iterative design process, we selected six task scenarios for further implementation of function-driven and expression-driven robot movements for a user study. This selection covers all four sectors of the representative space, comprising three function-oriented and three social-oriented tasks, as detailed in Section \ref{sec:study_material}.

\section{User Study}
\label{sec:study}

Our research question is whether movements driven by expressive utility can enhance users’ perceptions of the robot and their experience in human-robot interaction. To investigate this, we compare two robot conditions: one employs only function-driven movements ($\gamma = 0$ in Equation \ref{eq:utility}), while the other incorporates expression-driven movements in addition to function-driven ones, achieving the same goal states but through different trajectories ($\gamma > 0$). Our objective is to determine whether—and to what extent—incorporating expression design into the robot’s movements influences user interaction outcomes, and how these effects may vary according to the context of the tasks. 

\subsection{Research Questions and Hypothesis}

\textbf{RQ1}: To what extent does adding expression-driven movements, in addition to function-driven movements, influence users’ perceptions of the robot?


\textbf{H1}: Users will perceive a robot that combines expression-driven and function-driven movements as more engaging, human-like, and intelligent than one solely incorporates function-driven movements.

\textbf{RQ2}: Does the task context affect movement preferences?

\textbf{H2}: Users’ perceptions will vary by task context—expression-driven behaviors will be less favorable for function-oriented tasks and more favorable for social-oriented tasks.






\subsection{Method}
\label{sec:study_material}

We used a within-subject study design in which each participant viewed videos of the robot completing six different tasks, presented in a randomized order. After watching each video, participants rated their perception of the robot and its interaction with the human shown in the video. They were also encouraged to explain the reasoning behind their ratings, providing insights into which specific robot behaviors influenced their preferences.

To create the video demonstrations, a team of human-robot interaction researchers and animation designers iteratively designed and refined pre-recorded robot movement trajectories using the design primitatives proposed in section \ref{sec:primitives}. These trajectories were then implemented using the off-the-shelf WidowX arm controller to ensure smooth interactions. The videos used in this study are included in the supplementary materials.

We designed and implemented six scenarios, each presented in two conditions:

\textit{F}: A robot with function-driven movements only.

\textit{E}: A robot with both function-driven and expression-driven movements.

Details of the six task scenarios and robot movement description are provided below.


\paragraph{Photograph Light} 
Robot responds to user's hand gestures to move and offer desired lighting conditions for photography. \textit{F}: Move in response to user gesture and object position; \textit{E}: Move to express the curiosity toward the object by leaning forward, movements incorporating robot's attention to user command by looking back toward the user when detecting an instructive gesture.

\paragraph{Project Assistance} 
Robot observes user task and provides a corresponding video projection to guide the task. \textit{F}: Move to a target position for projection, and project a corresponding video; \textit{E}: Show curiosity toward the user activity and display joint attention through gaze direction.

\paragraph{Failure Indication} 
User instructs a goal position for the robot which is out of reach, the robot displays the error message back to the user. \textit{F}: Attempt to move toward the goal direction, reach the limit, and verbally output the error; \textit{E}: Pause to display hesitation before moving, stretch the body to display efforts when reaching the limit, and look back at the user and shake head before sending a verbal reaction.

\paragraph{Remind Water} Robot interrupts use activity and sends out a reminder to drink a cup of water. \textit{F}: Move to point toward the water cup, light up, and send a verbal reminder; \textit{E}: Move to the goal pose described in \textit{F}, push the cup toward the user, and gaze toward the user before sending the verbal reminder.

\paragraph{Social Conversation} Robot takes the role as a social companion, engages with the user in a social conversation about daily activities. \textit{F}: Respond to user's speech verbally; \textit{E}: Use movements as nonverbal cues in accordance with verbal texts, including gazing at the user, pointing to refer to the object in context of speech, use kinesthetic gestures to display emotions of excitement (dancing movement) and sadness (lowering the head).

\paragraph{Play Music}  Robot plays music entertainment accompanying user's daily activities. \textit{F}: Play music with no movement; \textit{E}: Play music while perform dance movements, align the movement rhythm with music tempo.

\begin{figure*}[t]
    \includegraphics[width=\textwidth]{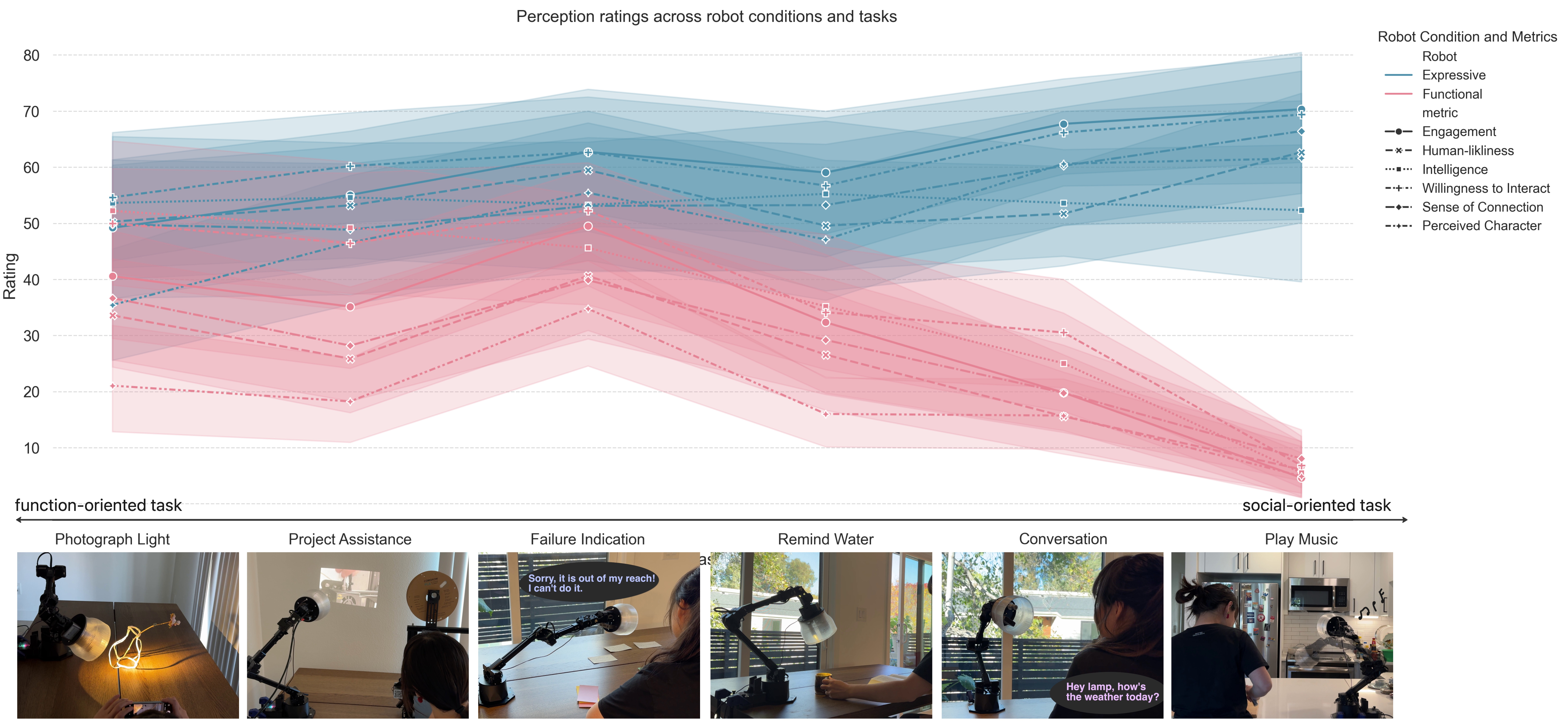}
     \caption{Quantitative Results: Comparing perception ratings between expression-driven (blue) and function-driven (pink) robot movements across six different task scenarios.}
     \label{fig:result}
\end{figure*}


\subsection{Measure}

We include six dimension of quantitative metrics to measure the perception toward the robot (\textit{human-likeliness, perceived intelligence, perceived emotion/character}), the quality of the interaction (\textit{interaction engagement, the sense of connection}), and \textit{willingness to use the robot} in real life. Specifically, participants rate their perception on a scale from 0 to 100 to indicate their agreement to six statements, measuring the above-mentioned aspects.
Besides, we collected demographic data of participants including their gender, age, background regarding robotics, background regarding expression design (such as performing arts, psychology, animation, communication), general level of empathy (\textit{``I find it easy to express empathy and understanding towards others.}), general acceptance toward robot (\textit{``I feel comfortable interacting with a robotic companion''}).
After each video, we collect qualitative feedback of the video by asking \textit{``how would you describe the robot in the video? What do you like or dislike about the robot?''} This allow us to gather insights on participants reasoning of their choices and explore open-ended ideas on the perception that we did not cover in the quantitative metrics.

\subsection{Participants}
We recruited 30 participants using emails and announcements distributed within the organization\footnote{The study is exempt under the organization's Human Study Review Board criteria. This study fits under the research involving benign behavioral interventions and collection of information from adults with their agreement (CFR 46. 104 (d) (3))}. Responses were filtered based on the time taken to complete the task, excluding those that took less than ten minutes, as well as any incomplete responses. This process resulted in 21 valid participants ($N=21$). Among them, eight are female, twelve are male, and one participant did not disclose their gender. The participants' ages range from 26 to 51 years. In terms of ethnicity, ten participants self-identified as Asian, nine self-identified as White or Caucasian, and two preferred not to disclose their ethnicity.

\section{Results}

This section presents the results through both quantitative and qualitative lens to uncover the perception differences between the two robot conditions across different tasks. 

\subsection{Quantitative Results}

To test H1, we compared the average ratings across different metrics between the two robot conditions, averaged across different tasks. 
The \textbf{robots with expression-driven movements are rated much higher} ($M=56.16, std=27.15$) \textbf{than robots incorporating only function-driven movements} ($M=28.77, std=27.15$). Welch's t-test revealed a statistical significance in the difference, $t=19.85, p<0.0001$.
The biggest difference lies in the metrics of Perceived character ($t=10.58$), followed by Human-likeliness ($t=9.32$), Engagement ($t=8.80$), Sense of Connection ($t=8.50$), then Willingness to Interact ($t=7.37$), and Perceived Intelligence ($t=5.22$), $p<0.001$ for all of the individual metric, which indicate statistical significance in the differences. Thus, \textbf{H1 is supported}.

Figure~\ref{fig:result} depicts the average ratings (from 0 to 100, the higher, the better) for expression-driven robots (blue) and function-driven robots (pink) across various tasks and evaluation metrics. The x-axis represents different tasks, arranged based on the purpose of the task—ranging from function-oriented tasks, such as photography lighting, projecting information, or displaying error messages, to socially-oriented tasks, such as music entertainment, social conversation, and habit nudging. 
The results reveal that expression-driven robots (blue) outperform function-driven robots (pink) across most of the tasks. The trend indicates that \textbf{for social-oriented tasks} (toward the right side of the x-axis), \textbf{expression-driven robots are perceived significantly better compared to function-oriented tasks} (toward the left side).

\begin{figure}[t]
    \includegraphics[width=0.5\textwidth]{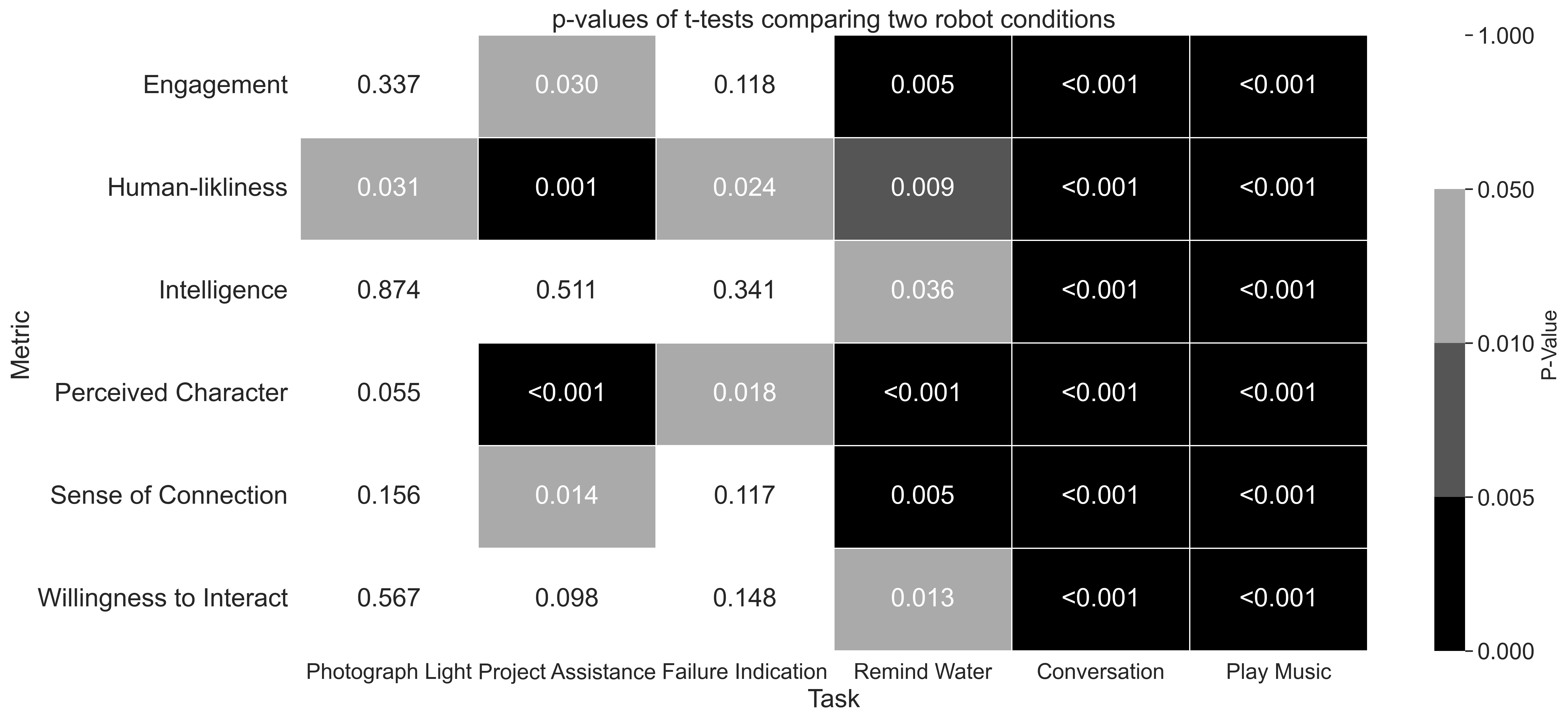}
     \caption{P-values of T-tests on user perception scores comparing expression-driven and function-driven movements.}
     \label{fig:pvalue}
\end{figure}

To further investigate these differences, we conducted statistical tests (Welch's t-test) for each task and metric to compare the expression-driven and function-driven robot movements. The resulting p-values are displayed in Figure~\ref{fig:pvalue}, where the dark color indicates statistical significance ($p < 0.05$). 
The table shows that for social-oriented tasks (\textit{play music, conversation, remind water}), expression-driven robots significantly outperform function-driven robots across all metrics. However, for function-oriented tasks (\textit{photograph light, project assistance, failure indication}), the two robots show no significant differences in metrics such as Perceived intelligence, Willingness to interact, and Engagement.
Thus, \textbf{H2 is supported}.

\begin{figure*}[t]
    \centering
    \includegraphics[width=\textwidth]{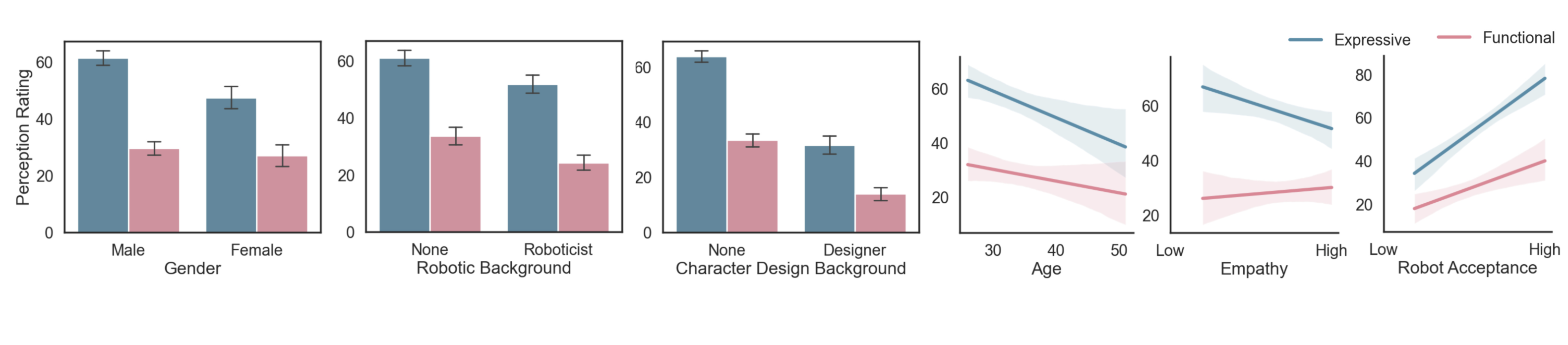}
     \caption{Effect of participants' demographics and backgrounds on average perception ratings}
     \label{fig:demographics}
\end{figure*}

To understand the effect of participants' backgrounds on their perception of robots, we conducted a linear regression analysis to examine the correlations between perception metrics (average ratings of perception) and background variables, including gender, age, general level of empathy, general acceptance of robots, backgrounds related to robotics and character design, as shown in Figure~\ref{fig:demographics}.
Our findings indicate that age significantly influenced perceptions of expressive robots, with older participants showing less preference for expressive robots ($p<0.001$). Additionally, we observed a trend in empathy levels affecting perception differences between functional and expressive robots: participants with self-rated low empathy perceived a stronger increase in robot likability after the integration of expressive movements. In contrast, participants who self-rated as having high empathy were less influenced by the integration of expressive movement in the robot.
We also found a positive correlation between robot acceptance and perception scores. However, these correlations did not reach statistical significance.
Besides, we conducted t-tests to compare the perception difference between gender groups, robotic backgrounds, and groups who have or do not have backgrounds related to character and expression design, including animation, psychology, performing arts, etc. Gender did not have a significant impact on perception ($p=0.2$). Robotic background is a significant predictor of perception, with non-roboticist rating robots higher than roboticist ($p=0.006$). Background related to expressive character design is another strong predictor, with experienced character designers and artists rating robots significantly lower than others. For all the groups above, they rated the expressive robots higher than the functional ones.

\subsection{Qualitative Results}
\begin{figure*}[t]
\centering
    \includegraphics[width=0.86\textwidth]{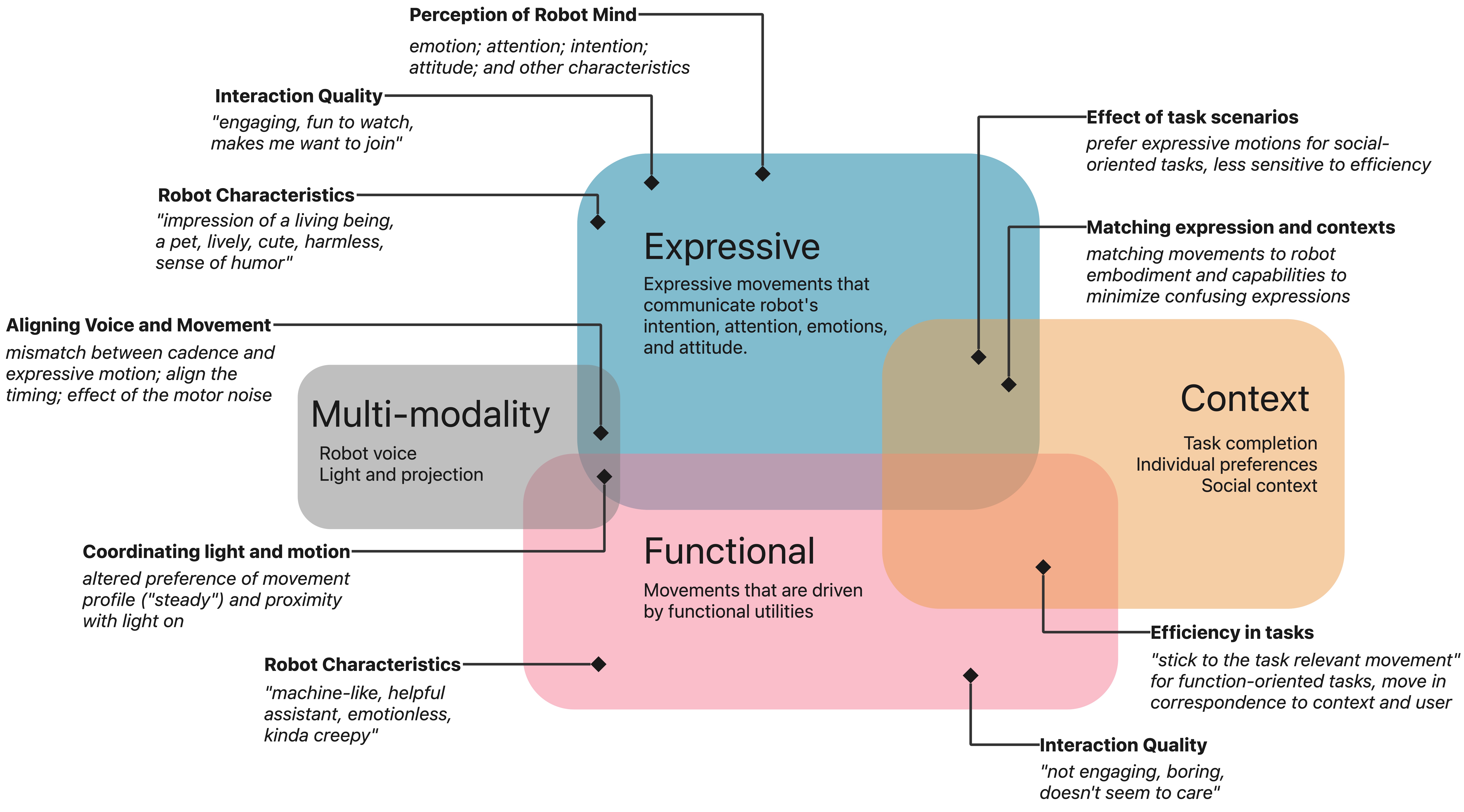}
     \caption{Summary of qualitative results - clusters and implications.}
     \label{fig:qualitative}
\end{figure*}

We conducted a qualitative thematic analysis of participants' feedback on individual robot behaviors to gain deeper insights into the reasoning behind their ratings. We identify the perception reasoning behind expressive and functional movements, as well as the interaction with task context and other interaction modalities, as visualized in figure \ref{fig:qualitative}.

\subsubsection{\textbf{Perception of Robot Characteristics}}
Participants commented the robot with \textbf{expressive movements} as more engaging, lively, harmless, embodied a \textit{``sense of humor''}, and were \textit{``fun to watch''}.
P4 noted that the expressive robot showed more information of robot internal states, such as emotions, which weren't apparent in function-driven only movements.
Several participants said the expressive motions were affective throughout the interaction of robot playing music and dancing:
\textit{``... that appeared to be matching her energy, dancing bigger because she was dancing bigger. \textbf{It makes me want to join.} There is such power in the synergy!'' (P12)}
In the qualitative reasoning, participants attributed the robot with expressive movements characteristics of human or pets, with its own drives and needs. Many participants said the expressive robot \textbf{remind them of a \textit{``puppy''} or \textit{``child''}}. In the failure indication scenario, P12 described the expressive robot has \textit{``a resilient spirit''}. In the playing music scenario, P1 said \textit{``it looks like it's having fun.''}

On the contrary, participants found it difficult to attribute human characteristics in robots with little movement and described the robot as a ``tool'' or compared it to existing home devices.
Participants commented the robot with only \textbf{functional movements} as \textit{``boring'', ``too machine-like'', ``not engaging'', ``\textbf{emotionless}''}, and may arouse negative feelings, especially during social conversations and playing music scenarios.
P3 is confused about its motivation of asking a social-oriented question during the conversation, as they assumed the robot did not have its own needs or emotions. 
P7 also commented that the robot looked \textit{``\textbf{kinda creepy} as if it was intently staring''} during the social conversation due to little movements.
Participants also noted the unnaturalness and lack of social connection with only functional movements.
During the scenario of reminding human to drink water, P1 commented that \textit{``It [the functional robot]\textbf{ does not seem to care} whether or not the human drinks the water.''} 
P15 mentioned that without the robot \textit{``looking at the user like how humans engage with each other, there was a lack of connection.''}

\subsubsection{\textbf{Inference of Robot State}}

Participants perceived and discussed the robot's intention, attention, attitudes and emotion behind the expressive movements in the questionnaires. For instance, in the social conversation scenario, P1 mapped the robot's movement toward the window as if \textit{``checking the weather outside'' }[\textbf{intention}], and the moving around as excitement, hanging its head as sadness [\textbf{emotions}]. 
During the projection task, P14 perceived the leaning forward and tilting head movement as robot's displaying curiosity, saying \textit{``The robot first seems to be interested in the human task [\textbf{attitude}]. I liked that. It seems to be happy to help.''}
In the task of photograph light, the expressive robot tilts the head back toward the the human when the human gestures. Several participants were able to perceive it as the robot paying \textbf{attention} to the human instruction. P7 said, \textit{``I liked when it looked at the person for feedback, as if saying "is this good?" ''}


Even for robots with \textbf{only functional movements}, some participants still perceive the robot's movement with theory of mind, such as projecting robot's attention and intention.
For example, during the task of failure indication, the robot stretches its arm toward the note before displaying the failure message, participants interpreted the stretch as \textit{``the robot seemed to struggle'' (P6)}. P7 also noted that the robot's facing direction clearly indicated its attention: \textit{``the robot and the person are looking at the same note''}.


\subsubsection{\textbf{What expressive movements were valued and what were not}}

While adding expressive movements proved to benefit the interactions, some participants found them unnecessary and \textbf{inefficient}. P15 noted for the failure indication task \textit{``... there needs to be a balance between engagement through motion and speed completion of the task being given, otherwise the human might grow \textbf{impatient}. It might be OK the first time with the novelty factor but will quickly fade out.''}
While some participants enjoyed the expressive motions, others noted that some expressive behaviors could be \textit{too exaggerated}, thus distracting or disturbing. Some participants mentioned they disliked the robot to move all the time, especially the movement for no apparent reasons, which may imply \textit{``a lack of attention on the robot's part'' (P5, conversation)}
Most participants appreciated the information that were \textit{``quick and easy to interpret''}, while for the subtle movements, participants had different preferences.

Participants reported negative perception when there was a \textbf{mismatch} between robot's movement and its perceived capabilities. For the failure indication function, P14 noted that \textit{``I did not like that it tried to get momentum with an impulse as it seems fake.''} P20 thought the robot did not have a camera on the head thus the ``looking at notes'' action seemed fake.

The preferences of adding expressive movements \textbf{vary across tasks}. For tasks that include little functional movements, and for tasks that were social-oriented and less sensitive to efficiency, expressive movements were more appreciated. For instance, in the scenario of playing music for entertainment, P21 noted that \textit{``I really liked this application for robot engagement! No fast responses were necessary, so having an engaging dancing motion made the robot more engaging. ''}
On the other hand, for the tasks that inherently have clear function-driven movements and are more function-oriented, adding expressive movements could be confusing to some individuals and preferences vary.
For instance, in the scenario of photograph light, participants thought the expressive movements made the robot seem less ``predictable''(P5), less ``steady''(P14). P18 wished the robot to \textit{``stick to only the task relevant motion which is angle and the light''}. Even without expressive movements, in such function-oriented tasks, participants rated highly of the robots as long as the robot were able to move in correspondence to the context and user request. 

It is important to note that many participants were less acceptable to robot taking proactive roles than reactive roles, such as reminding the user to drink water. For instance, P20 noted that \textit{``... I don't like my life to be controlled by a robot. If I'm in the middle of some exciting readings, I don't want to be disrupted by a robot's command.''}
Adding the expressive movements such as with a playful character could increase the acceptance of robot behavior, P8 noted that \textit{``\textbf{Without the playfulness, I might find this type of interaction with a robot annoying rather than welcome and engaging.}''}

\subsubsection{\textbf{The effect of voice and light}}
Participants repeatedly commented on the alignment between movements and other modalities, such as the robot's sound and light.
Several participants felt there was a \textbf{mismatch} between robot's speech \textbf{cadence} and the expressive motions - while the expressive motions were \textit{``endearing, showing a character''}, the tone from the robot was very \textit{``automated'', ``stiff''}, and \textit{``took away from how friendly the interaction felt''.}
P7 and P15 noted that the \textbf{timing of the voice} need to align with the timing of the movement, to make it feel more natural. P12 found the sound from the motors disturbing, and may only \textit{``punctuate with smaller slower movements''.}

The coordination between movement and \textbf{lighting} can influence the comfort of interaction. Some participants mentioned preferring the robot to \textbf{remain steady} while maintaining the light. In such cases, expressive movements might interfere with the primary lighting function, as the robot's motion could distract when displaying attention or curiosity.
Additionally, the proximity of the light also impacted perceptions of disruption, as noted by P21.
P7 appreciated that \textit{``the robot turned out its light when looking at the person''} during assistive projection.

\section{Discussion}

In this paper, we conducted design research to explore how adding expressive movements on top of purely functional ones affects human-robot interactions. Both quantitative and qualitative findings show that expressive movements, compared to strictly functional ones, enhance the overall interaction experience and improve perceived robot qualities. Participants were more likely to recognize the robot’s state of mind, projecting intentions, attention, emotions, and attitudes throughout their interactions. For instance, participants recognized the robot’s “gaze” as a marker of joint attention, suggesting a stronger bond between human and robot.
Additionally, participants more frequently described the expressive robot as a living being—for example, a “pet,” a “child,” or a “friend.” In some tasks, adding expressive movements made the experience more engaging and playful. In particular, when the robot initiated an interaction or nudged participants, its expressive movements made those interruptions feel more acceptable, such as in the case of interrupting the user during the reading and nudging the user to drink water. This may be due to that participants possess more empathy towards the robot with expressive movements, as they remind them of living beings, just like the pets making the mess in the home; and thus the initially disturbing behaviors transfer into a playful social interaction. This highlighted the benefit of adding expressive and characterful motions for robot-initiated task scenarios.

The quantitative results reveal differences in perception across various tasks. For social-oriented tasks—such as playing music, engaging in social conversations, and nudging water—adding expressive movements was significantly more preferred. The qualitative reasoning further illuminates this trend: in these tasks, users prioritize engagement and entertainment over task efficiency. Consequently, adding expressive movements enhances the robot’s playfulness and character.
Moreover, social-oriented tasks in the study generally entail minimal function-driven motions. For instance, when playing music or holding social conversations, the robot primarily responds verbally, and the functional output does not involve any physical movement. In these contexts, incorporating expressive movements aligned with the social and task setting enriches the interaction, increases user engagement, and can even convey an additional layer of information.
On the other hand, for tasks that are function-driven—such as adjusting lighting angles or shifting between projection spaces—adding expressive motions can disrupt the robot’s primary function and potentially cause confusion or annoyance for users.
This implies that expression-driven movements need to complement function-driven movements by adjusting both the amount and timing of expressions to enrich—rather than conflict with—the original motions. Future research should balance the trade-off between task efficiency and characterfulness in human-robot interaction, while also considering individual preferences through personalized behaviors. For instance, although some users enjoyed a more animated robot, others disliked constant movement, particularly when it occurred without a clear or explicit reason.

The design and integration of expressive motions also need to align with the robot’s embodiment and capabilities. For instance, gaze behaviors should be co-designed with the placement of the robot’s ``eyes'' (cameras) and ``head''. While this may be intuitive for humanoid robots, robots with non-anthropomorphic features rely on appearance design and movement patterns to suggest a life-like embodiment that can be intuitive to humans and even other species.
It is equally important to match these movements with the robot’s other modalities—in this case, its voice and the lamp’s light or display. As many participants noted, the speech content, tone, and timing during movement sequences all play a key role in shaping the perceived quality of the robot’s behaviors. Future research needs to consider more extensive alignment among these different modalities to further enhance human-robot interaction.

\section{Conclusion}

In this paper, we present \textbf{ELEGNT}, a framework for designing expressive and functional movements for non-anthropomorphic robots in daily interactions. The framework integrates function-driven and expression-driven utilities, where the former focuses on finding an optimal path to achieve a physical goal state, and the latter motivates the robot to take paths that convey its internal states—such as intention, attention, attitude, and emotion—during human-robot interactions.
We use a lamp-shaped robot to illustrate the design space for functional and expressive movements in various interaction scenarios, ranging from function-oriented to social-oriented tasks, and involving reactive versus proactive robot roles.
We conduct a user study to compare perceptions of the robot when using expressive movements versus only functional movements across six different task scenarios. Our results indicate that incorporating expressive movements significantly increases user likability toward the robot and enhances interaction engagement. The perception varies across tasks, with social-oriented tasks that require minimal function-driven movements benefiting particularly from the addition of expression-driven movements.
Qualitative analysis further elaborates on users' differing perceptions of the robot's characteristics and the perceived robots' mental models. The findings also highlight the importance of aligning movement with other robot modalities, such as voice and light.
Future work will integrate these design insights into a generative framework for creating context-aware robotic movements that effectively express intentions in non-anthropomorphic robots.



\bibliographystyle{ACM-Reference-Format}
\bibliography{main}

\end{document}